\documentclass[journal,twoside,web]{ieeecolor}

\usepackage{lcsys}
\usepackage{cite}
\usepackage{amsmath,amssymb,amsfonts,amscd,dsfont}
\usepackage{mathtools}
\usepackage{accents} 
\usepackage{algorithm, algorithmic, listings}
\usepackage{graphicx,tabularx,adjustbox}
\usepackage{textcomp}

\usepackage[caption=false]{subfig}
\usepackage{booktabs}
\usepackage{enumerate}

\DeclareMathOperator*{\argmax}{arg\,max}
\DeclareMathOperator*{\argmin}{arg\,min}

\newtheorem{proposition}{Proposition}

\newtheorem{definition}{Definition}

\newtheorem{problem}{Problem}

\usepackage{multirow}

\def\BibTeX{{\rm B\kern-.05em{\sc i\kern-.025em b}\kern-.08em
    T\kern-.1667em\lower.7ex\hbox{E}\kern-.125emX}}
\begin{document}
\title{A Value Function Space Approach for Hierarchical Planning with Signal Temporal Logic Tasks}
\author{Peiran Liu*, Yiting He*, Yihao Qin, Hang Zhou and Yiding Ji$^{\nmid}$, \IEEEmembership{Member, IEEE}
\thanks{*Equally contributed, $\nmid$ corresponding author}
\thanks{All authors are with Robotics and Autonomous Systems Thrust, Systems Hub, Hong Kong University of Science and Technology (Guangzhou), Guangzhou, China. (Emails: \{pliu868, yhe398, yqin637, hzhou269\}@connect.hkust-gz.edu.cn, jiyiding@hkust-gz.edu.cn)}
\thanks{This work is supported by National Natural Science Foundation of China grants 62303389,  62373289; Guangdong Basic and Applied Basic Research Funding grants 2022A151511076, 2024A1515012586; Guangdong Scientific Research Platform and Project Scheme grant 2024KTSCX039; Guangzhou-HKUST(GZ) Joint Funding Program grants 2024A03J0618, 2024A03J0680.}}

\pagestyle{empty} 

\maketitle

\thispagestyle{empty} 

\begin{abstract}
Signal Temporal Logic (STL) has emerged as an expressive language for reasoning intricate planning objectives. However, existing STL-based methods often assume full observation and known dynamics, which imposes constraints on real-world applications. To address this challenge, we propose a hierarchical planning framework that starts by constructing the Value Function Space (VFS) for state and action abstraction, which embeds functional information about affordances of the low-level skills. Subsequently, we utilize a neural network to approximate the dynamics in the VFS and employ sampling based optimization  to synthesize high-level skill sequences that maximize the robustness measure of the given STL tasks in the VFS. Then those skills are executed in the low-level environment. Empirical evaluations in the Safety Gym and ManiSkill environments demonstrate that our method accomplish the STL tasks without further training in the low-level environments, substantially reducing the training burdens.
\end{abstract}

\section{Introduction}

\IEEEPARstart{C}{ontrolling} robots and autonomous systems to accomplish long-horizon, safety-critical and time-sensitive tasks is intrinsically challenging, especially when it comes to rigorously reason about the dynamic behaviors and provide provable guarantees. Formal methods originate from software engineering and empower the formulation of structured specifications through formal languages~\cite{belta2017formal}. In recent years, formal methods have been employed in a wide range of applications, e.g., robot planning~\cite{li2019formal}, autonomous vehicles~\cite{huang2023formal}, multi-agent systems~\cite{liu2025controller}, smart grids~\cite{beg2018signal} and industrial automation~\cite{zhou2023local}.

STL is an expressive formal language for specifying complex tasks, encompassing both quantitative and qualitative properties \cite{lindemann2025formal}. STL-based planning and control methods utilize various optimization techniques. For example, \cite{sun2022multi, kurtz2022mixed} employ Mixed-integer Linear Programming (MILP), and \cite{takayama2025stlccp} uses convex optimization. \cite{haghighi2019control} adopts a smooth robustness measure for nonlinear programming (NLP), representing classical optimization methods. For more complex systems, \cite{leung2023backpropagation} applies smooth gradient techniques to neural network backpropagation. Additionally, \cite{singh2023stl, wang2024tractable, wang2024synthesis} use reinforcement learning (RL) for policy development. These methods focus on solving one particular temporal logic specification.

Meta-RL, on the other hand, aims to learn a policy that adapts to new tasks and requires limited additional training steps. The work~\cite{vaezipoor2021ltl2action} leverages Graph Neural Network (GNN) to encode and adapt to new LTL instructions. Similarly,~\cite{yalcinkaya2024compositional} proposed representing tasks using Deterministic Finite Automata (DFA) and learning embeddings for these automata-represented tasks. Furthermore,~\cite{araki2021logical} utilizes Finite State Automata (FSA) to represent LTL, with the goal of combining subpolicies for new LTL tasks with minimal retraining steps. While these methods convert LTL tasks into FSA representations, we aim to explore such capability for STL tasks.

Inspired by VFS ~\cite{alexander2022value}, this work proposes a STL guided skill planning framework. Our approach is hierarchical and generates sequences of skills to satisfy STL formulas without additional low-level skill training, which effectively decouples high-level planning in the VFS from low-level dynamics in the original environment. That is, we abstract the planning space to a more succinct VFS. Our method constructs the VFS from value functions of RL and employs neural networks to approximate VFS dynamics. We reuse skills in the VFS to avoid training at low-level Markov Decision Process (MDP)  for new STL tasks. Then we apply sampling-based techniques to generate the optimal skill sequence to achieve the maximum STL robustness scores in the VFS, which are subsequently executed in the MDP space. Simulations conducted in the Safety Gym and ManiSkill environments validate the performance of our method by showing that accomplishing STL tasks in the VFS ensures their satisfaction in the low-level state space. 

The remainder of the work is organized as follows. Section~\ref{sec:preliminary} introduces the preliminary knowledge of STL and RL, then formulates the STL planning problem. Section~\ref{sec:method} constructs the VFS and proposes a VFS based skill planning framework. Section~\ref{sec:experiment} includes the simulation results to demonstrate the performance of our method in robot task planning scenarios. Finally, Section~\ref{sec:conclusion} concludes the work and proposes several potential extensions. 
\vspace{-8pt}
\section{Preliminaries and Problem Formulation}\label{sec:preliminary}

STL describes dynamic behaviors by real value signals $ \mathbf{s} = s_0, s_1, ..., s_T$ where $s_i \in \mathbb{R}^n$ and \(\left(\mathbf{s}, [t_1, t_2]\right)\) stands for $s$ in time interval $[t_1, t_2]$ with \(t_1, t_2 \in \mathbb{N}\) and \(t_2 \geq t_1\)~\cite{deshmukh2017robust}. STL syntax comprises three elements: predicates \(\mu\), boolean operators \(\wedge\) and \(\neg\), and temporal operators constrained by time intervals \(\mathbf{U}_{[t_1, t_2]}\). A STL formula $\phi$ is defined recursively as: 
\vspace{-3pt}
\begin{equation}
    \phi::= \text{True} \ | \ \mu \ | \ \neg \phi \ | \ \phi_1 \land \phi_2 \ | \ \phi_1 \mathbf{U}_{[t_1,t_2]}\phi_2
    \label{eq:stl_formula}
\end{equation}
where $\mu: \mathbb{R}^n \rightarrow \{ \text{True, False}\}$ is an atomic predicate that assigns a boolean value to a signal, typically in the form of \(\mu(\mathbf{s}) \geq 0 \). The ``\emph{until}" operator \(\phi_1 \mathbf{U}_{[t_1,t_2]}\phi_2\) holds when \(\phi_{1}\) remains true until \(\phi_{2}\) becomes true. Additional temporal operators are derived from the above syntax, e.g., ``\emph{eventually}" is defined as \(\mathbf{F}_{[t_1, t_2]} \phi := \text{True} \mathbf{U}_{[t_1, t_2]} \phi\), and ``\emph{globally}" is defined as \(\mathbf{G}_{[t_1, t_2]} \phi := \neg(\mathbf{F}_{[t_1, t_2]} \neg \phi)\). We write \((\mathbf{s}, t) \models \phi\) if from some time instant \(t\), signal \(\mathbf{s}\) satisfies \(\phi\).

\begin{definition}[Robustness of STL]
Let $\phi$ be an STL formula and $\mathbf{s} = s_0, s_1, ..., s_T$ be a signal, the robustness score of $\phi$ w.r.t $\mathbf{s}$ at time $t$, denoted by $\rho(\phi, \mathbf{s}, t)$, is defined as:

\vspace{-10pt}
\begin{equation}
    \begin{aligned}
        & \rho(\mathbf{s},t, \top)= 1 , \quad \rho(\mathbf{s},t, \mu\geq 0)= \mu(\mathbf{s}(t))\\
        & \rho(\mathbf{s},t, \neg \phi)= -\rho(\mathbf{s},t,\phi)\\
        & \rho(\mathbf{s},t, \phi_1 \land \phi_2)= \min\{\rho(\mathbf{s},t,\phi_1),\rho(\mathbf{s},t,\phi_2) \}\\
        & \rho(\mathbf{s},t, \phi_1 \lor \phi_2)= \max\{\rho(\mathbf{s},t,\phi_1),\rho(\mathbf{s},t,\phi_2) \}\\
        & \rho(\mathbf{s},t, \phi_1 \implies \phi_2)= \max\{-\rho(\mathbf{s},t,\phi_1),\rho(\mathbf{s},t,\phi_2) \}\\
        & \rho(\mathbf{s},t, \phi_1 \mathbf{U}_{[t_1,t_2]} \phi_2)= \\
        & \quad \sup\limits_{t'\in[t+t_1,t+t_2]}\min\left\{\rho(\mathbf{s},t',\phi_2), \inf\limits_{t''\in[t,t']}\rho(\mathbf{s},t'',\phi_1) \right\} \\
        & \rho(\mathbf{s},t, \mathbf{F}_{[t_1,t_2]} \phi)= \sup\limits_{t'\in[t+t_1,t+t_2]}\rho(\mathbf{s},t',\phi)\\
        & \rho(\mathbf{s},t, \mathbf{G}_{[t_1,t_2]} \phi)= \inf\limits_{t'\in[t+t_1,t+t_2]}\rho(\mathbf{s},t',\phi)
    \end{aligned}
    \label{def:robustness_score}
\end{equation}
\end{definition}

The value \(\rho(\mathbf{s}, \phi, t)\) quantifies how well a signal \(\mathbf{s}\) satisfies a formula \(\phi\) at time \(t\). Several methods exist to compute the score and we adopt the robustness metric method in~\cite{deshmukh2017robust}.

The environment is formally modeled as a deterministic MDP denoted by $ M = (S, A, f, R) $ where $ S $ denotes the finite state space, $ A $ is the finite action space, $ f: S \times A \rightarrow S $ is the deterministic transition function and $ R: S \times A \rightarrow \mathbb{R}$ is the reward function.
The MDP states evolve under a sequence of actions $ \{a_t\}_{t=0}^{T-1}=(a_0, a_1,\ldots, a_{T-1})$, producing a trajectory $s_{0:T} = (s_0, f(s_0, a_0), \dots, f(s_{T-1}, a_{T-1}))$ with discounted accumulative return $ \sum_{t=0}^{T} \gamma^t R(s_t,a_t) $. The objective is synthesizing a control policy $\pi$ that maximizes the discounted accumulative return. Since the dynamics of MDP is unknown, RL is employed for optimal decision making based on the information of states, actions and rewards~\cite{bertsekas2025course}. Two key outcomes of RL algorithms are the optimal value function $ V^*: S \rightarrow \mathbb{R} $ and the optimal policy $ \pi^*: S \rightarrow A $, which should satisfy their respective Bellman optimality equations:

\vspace{-8pt}
\begin{equation}
    V^*(s) = \max_{a \in A} [ R(s,a) + \gamma V^*(f(s,a)) ] 
    \label{eq:value function}
    \vspace{-5pt}
\end{equation}
\vspace{-5pt}
\begin{equation}
    \pi^*(s) = \argmax_{a \in A} [ R(s,a) + \gamma V^*(f(s,a)) ]
    \label{eq:optimal policy}
\end{equation}
\vspace{-13pt}

\begin{problem}[STL guided planning]\label{prob:state_optimization}
Given an MDP $ M = (S, A, f, R) $ with unknown dynamics $f$ and a STL formula $\phi$, the goal is to synthesize an optimal action sequence of $T\in \mathbb N^+$ time steps to maximize the robustness score of $\phi$:
    \vspace{-3pt}
    \begin{equation}
     a^*_0, a^*_1,\ldots, a^*_{T-1}=\argmax_{a_0, a_1,\ldots, a_{T-1}} \rho(s_{0:T}, \phi)
    \vspace{-5pt}
    \end{equation}
where $s_{0:T} = (s_0, f(s_0, a_0), \dots, f(s_{T-1}, a_{T-1}))$ is a sequence of states, recursively generated by selected action sequence $(a_0, a_1, ..., a_{T-1})$ under the transition function $f$.
\end{problem}

The MDP state space is usually prohibitively high-dimensional due to extensive robot sensor data, which poses challenges for directly tacking the problem. We will develop an abstraction based solution to reduce the dimension of the state space, as detailed in the following section.
\vspace{-3pt}
\section{Signal temporal logic guided skill planning}\label{sec:method}

\begin{figure}[ht!]
    \centering
    \includegraphics[width=9cm, height=7.2cm]{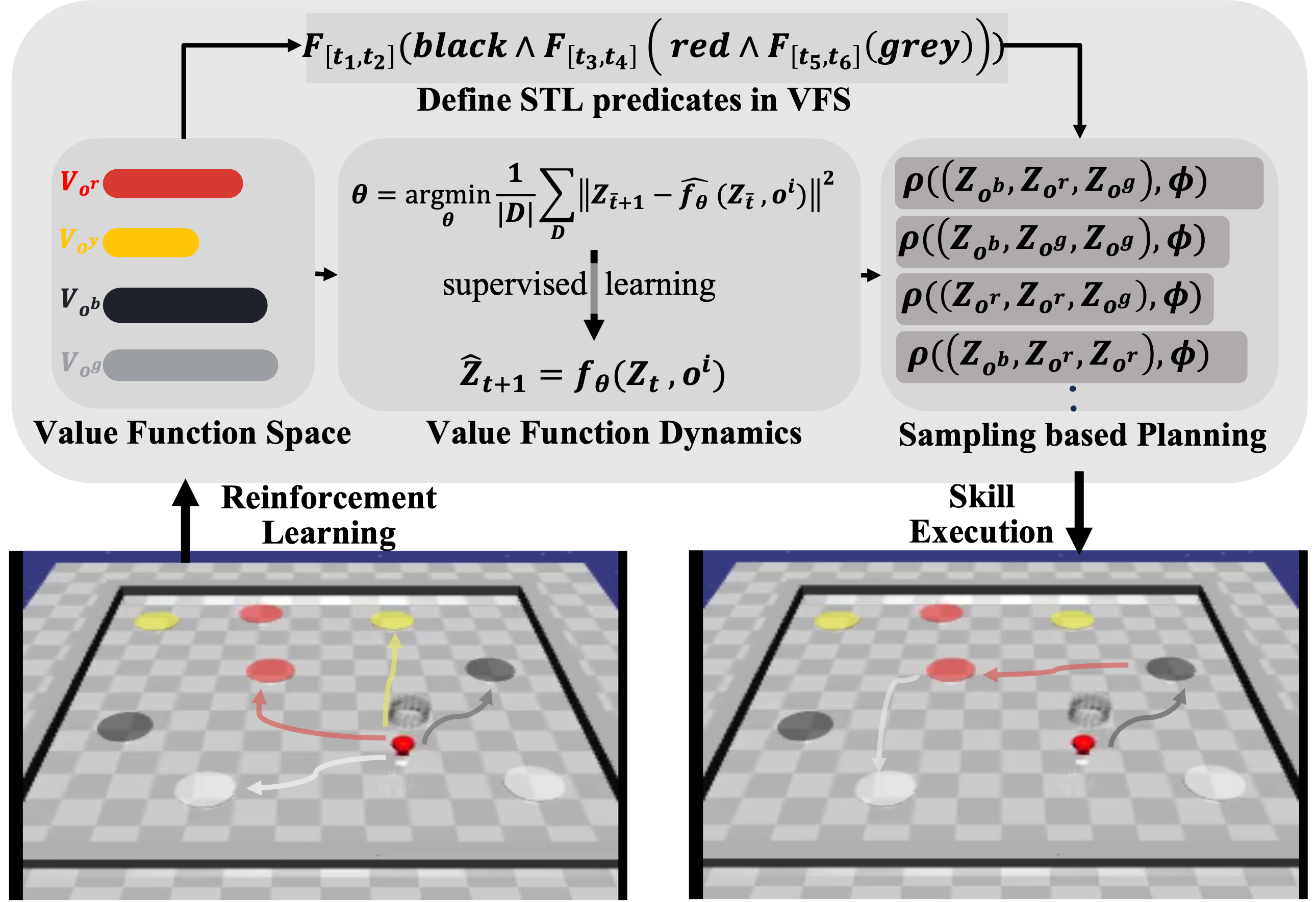}
    \vspace{-20pt}
    \caption{\footnotesize Hierarchical framework of STL guided planning}
    \label{fig:pipeline}
    \vspace{-15pt}
\end{figure}

In this section, we develop a hierarchical STL planning framework outlined in Fig.\ref{fig:pipeline}. First, we define goals of RL and train skills from value-based RL with sparse rewards. Next, VFS is constructed as an abstraction of the original state space, which captures the interactions between the agents and the environment. Supervised learning is employed to approximate the transition dynamics in VFS. Then we define both reach and avoid predicates for the given STL tasks, followed by a reformulation of Problem~\ref{prob:state_optimization} in VFS. Finally, we employ the random sampling to generate an optimal skill sequence to provably complete the STL task in VFS, which also significantly reduces the computational cost. 

\vspace{-10pt}

\vspace{-8pt}
\subsection{Skills learned with value-based RL}\label{subsec:goalskill}

In our framework, a skill is defined analogously to an option in the Options framework of ~\cite{bertsekas2025course}. Given an MDP $M=(S, A, f, R)$, each skill $o$ is composed of three components: a policy $\pi: S \rightarrow A$, a termination condition $\beta: S \rightarrow \{0, 1\}$, and an initiation set $L \subseteq S$. Specifically, a policy $\pi$ is derived using value-based RL and the termination condition is evaluated based on STL predictions, as later detailed in Section \ref{sec:method}.C. For simplicity, we assume that all MDP states are included in the initiation set, i.e., $L = S$.

To generate skills from MDP, we first define a \emph{goal-augmented MDP} as $M_g = (S, A, f, R, G)$, where $S$, $A$, $f$, $R$ remain the same as MDP and $G = \{g_1, g_2, ..., g_k\}\subseteq S$ is the set of goal regions, with $g_i \in S$ being a specific state of $S$. Inspired by~\cite{qiu2024instructing}, we use reinforcement learning to acquire skills that guide the agent towards reaching the goals.

Any RL algorithm suffices as long as it successfully learns the value function in Equation (\ref{eq:value function}) with a sparse reward, and we use Proximal Policy Optimization (PPO)~\cite{bertsekas2025course}. Given a goal $g_i\in G$, its associated sparse reward function is defined as: $r_{g_i}=\mathbb I(g_i~\text{is satisfied})$ where $\mathbb I$ is the indicator function, that is, $r_{g_i}=1$ only when $g_i$ is achieved, otherwise $r_{g_i}=0$. 

Then we leverage the concept of ranking function from \cite{urban2016synthesizing} and show that the optimal value function is a special ranking function in our context of RL towards reaching goals.

\begin{definition}[Increasing ranking Function]\label{def:ranking function}
    Given an MDP $M=(S, A, f, R)$, a increasing ranking function $\xi: S \rightarrow \mathbb{R}^+$  (i) increases through transitions, i.e., $\forall s, s'\in S$, $\forall a\in A$: $s' = f(s, a)\Rightarrow \xi(s') \geq \xi(s)$; (ii) is bounded from above, where $\forall s \in S: \xi(s) \leq 1$; (iii) has the upper bound at terminal states, i.e., $\xi(s)=1$ if $s\in S$ is a terminal state.
\end{definition}

Simple tasks with proper reward functions render it possible to learn optimal policies reaching terminal states. Ranking functions measures the level of task completeness.

\begin{proposition}\label{prop:increasing ranking function}
Given a goal-augmented MDP $M_g = (S, A, f, R, G)$ and a goal $g\in G$, the optimal value function $V^*$ learned with the above sparse reward is an increasing ranking function since it (i) increases by transitions under the optimal policy, i.e., $\forall s, s'\in S$: $s' = f(s, a^*) \Rightarrow V^*(s') \geq V^*(s)$ where $a^* = \pi^*(s)$ is the optimal action in Equation (\ref{eq:optimal policy}); (ii) is upper bounded where $V^*(s) \leq 1, \forall s \in S$; (iii) hits the upper bound at the goal, i.e., $V^*(s) = 1 \Leftrightarrow s=g$.
\end{proposition}

\begin{proof}
\label{proof: increasing ranking function}
In the above mentioned context of RL with sparse rewards, when a state-action pair $(s, a_g)$ leads a transition to a terminal state, i.e., $f(s, a_g) = g$, the reward is set to $1$, i.e., $R(s, a_g) = R(g) = 1$.  Conversely, if the state-action pair $(s, a)$ leads to a non-terminal state, the reward is set to $0$, i.e., $R(s, a) = R(s) = 0$. (iii) when goal regions are reached, we have $V^*(g) = R(g) = 1$ for $g\in G$. (i) By Bellman equation $V^*(s) = \max_a (R(s, a) + \gamma V^*(f(s, a)))$, we have $V^*(s) = R(s) + \gamma V^*(f(s, a^*)) $ where for $s \notin G$, $\gamma \in [0, 1]$ and $R(s)=0$ imply that $V^*(s) \leq V^*(f(s, a^*)) = V^*(s')$. (ii) there are two cases: $s = g$ and $V^*(s) = 1$; or $s \neq g$ and $s$ is one step from $g$, that is $g = f(s, a^*)$, which means $V^*(s) \leq V^*(g) = 1$. Thus, $\forall s\in S$, $V^*(s)\leq 1$ holds. 
\end{proof}

Proposition \ref{prop:increasing ranking function} implies that $V^*$ reflects the ``distance" between the goal state and the current state. $V^*(s) = 1$ indicates that the current state is the goal state, while the agent avoids the goal by taking actions leading to $V^*(s) < 1$. This observation will play a role in handling STL tasks. 

\vspace{-8pt}
\subsection{Construct Value Function Space}\label{subsec:vfs}

Suppose that we have $k$ goals ($|G| = k$), and will train $k$ skills $o^i \in O$ for each $g_i \in G$ through RL with rewards defined in the last subsection. This process also returns the respective skill value functions $V_{o^i}$ to facilitate the construction of an embedding space $Z$ to abstract the MDP environment, which maps a state $s_t$ to a $k$-dimensional vector $Z(s_t) \equiv [V_{o^1}(s_t), V_{o^2}(s_t), \ldots, V_{o^k}(s_t)]^T$ called a VFS ~\cite{alexander2022value}. Through high-level skill execution, the VFS effectively captures functional information about potential interactions between the agent and the environment, thereby being a scalable abstraction of the low-level MDP.

Note that the time steps in the high-level VFS differ from their counterparts in the low-level MDP. To distinguish between them, we use $t$ to represent the MDP time index, $T$ to denote the total number of low-level steps, $\bar{t}$ to indicate the VFS time index, and $\bar{T}$ for the total VFS steps. A single high-level VFS time step from $\bar{t}$ to $\bar{t}+1$ consists of $\tau$ time steps in the low-level MDP. Consequently, the total time horizon in the original MDP state space is $T = \tau \cdot \bar T$.

Given a goal set $G=\{g_1, g_2, \ldots, g_k\}$, a set of skills $O = \{o^1, ..., o^k\}$ for each $g_i$ and an arbitrary STL formula $\phi$ constructed with respect to $G$ (e.g. $\phi = \mathbf{F}_{[0, T]} g_1 \land \mathbf{G}_{[0, T]} \neg g_2$, which represents eventually reaching $g_1$ and always avoiding $g_2$ within time interval $[0, T]$), we reformulate Problem \ref{prob:state_optimization} as:

\begin{problem}[STL guided skill planning in VFS]\label{prob:skill_optimization}
Given a goal-augmented MDP $ M = (S, A, f, R, G) $ with unknown dynamics $f$, a STL formula $\phi$ and a skill set $O=\{o^i, o^j, \ldots, o^k\}$, we aim to synthesize an optimal skill sequence for $\bar T$ high-level steps to maximize the robustness score:

\vspace{-12pt}
\begin{equation}
    o^*_0, o^*_1, \dots, o^*_{\bar T-1} = \argmax_{o_0, o_1, \dots, o_{\bar T-1}} \rho(Z_{0:\bar T}, \phi)
    \vspace{-5pt}
\end{equation}
where $Z_{0:\bar T} = (Z_0, Z_1, \dots, Z_{\bar T})$ is a sequence in VFS.
\end{problem}

To solve Problem~\ref{prob:skill_optimization}, we still need to learn the transition dynamics and define STL predicates in the VFS. For this purpose, we first approximate the transition dynamics in the VFS, denoted by $\hat{f}_\theta: Z \times O \rightarrow Z$, such that $Z_{\bar{t}+1} = \hat{f}_\theta(Z_{\bar{t}}, o^i)$ for all $o^i \in O$. This is achieved through supervised learning, using a dataset of prior random interactions within the environment. The initial position and environment are randomly configured for data collection. The dataset $D = \{(Z_{\bar{t}}, Z_{\bar{t}+1}, o^i), \ldots\}$ is collected by observing the current VFS $Z_{\bar{t}} = Z(s_t)$, executing a random skill $o^i$ for $\tau$ steps, and subsequently observing the resulting VFS $Z_{\bar{t}+1} = Z(s_{t+\tau})$. The transition $\hat{f}_\theta$ operates on high-level skill steps, each encompassing $\tau$ steps of low-level actions. Specifically, when executing a high-level action using a skill $o^i$, transitioning from $Z_{\bar{t}}$ to $Z_{\bar{t}+1}$ in the state space involves executing $o^i$ for $\tau$ low-level action steps, resulting in a transition from state $s_t$ to state $s_{t+\tau}$. The parameter $\theta$ is calculated by minimizing the mean squared error over $D$: 

\vspace{-5pt}
\begin{equation}
    \theta = \argmin_\theta \frac{1}{|D|} \sum_D || Z_{\bar{t}+1} - \hat{f}_\theta(Z_{\bar{t}} ,o^i)||^2
\end{equation}
\vspace{-8pt}

\begin{figure}[ht!]
    \centering
    \includegraphics[width=8.5cm, height=4.5cm]{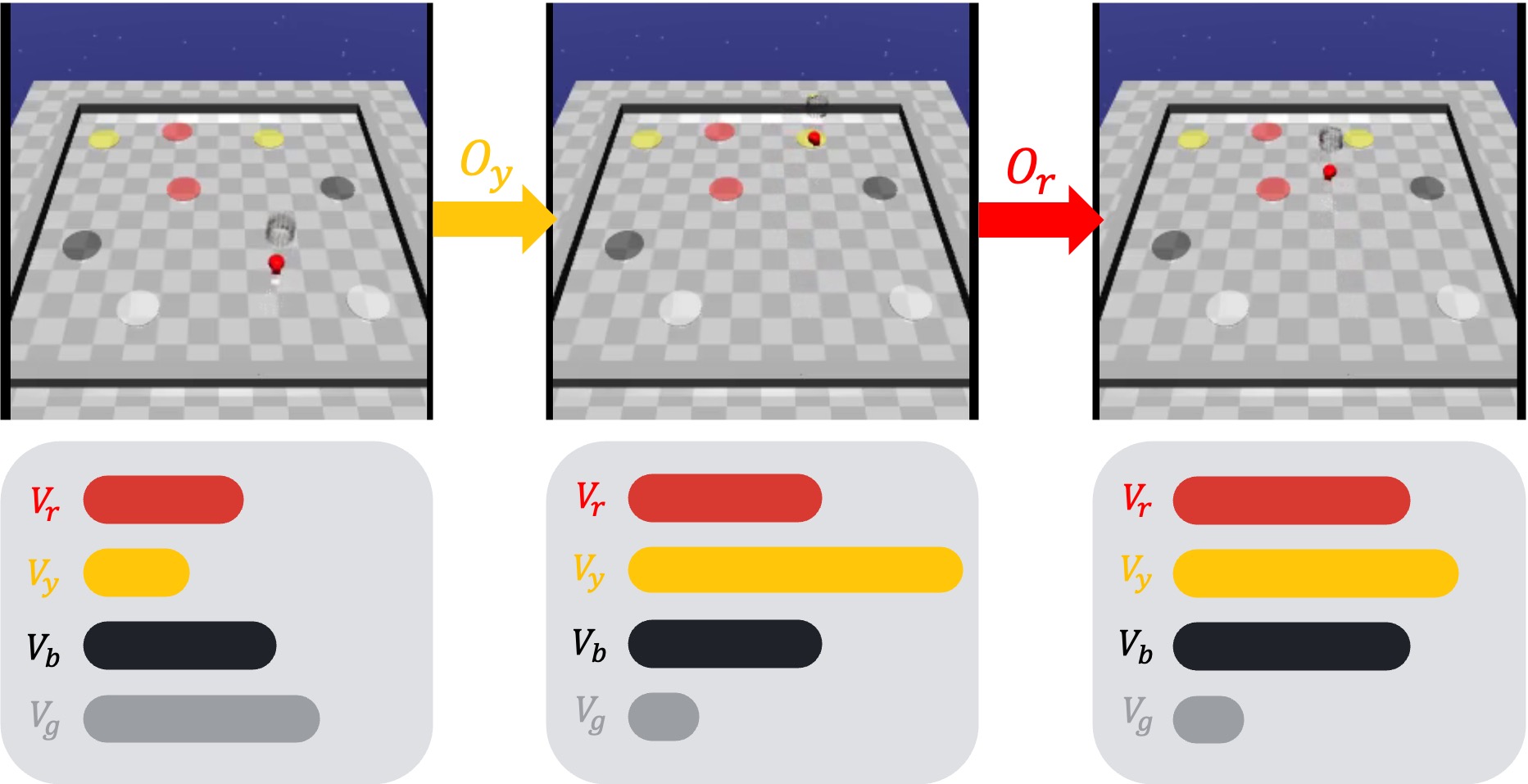}
    \vspace{-5pt}
    \caption{\footnotesize Top: trajectories in the original state space under execution of high level skills; bottom: value changes of skill value functions}
    \label{fig:VFDynamic}
    \vspace{-20pt}
\end{figure}


Fig.\ref{fig:VFDynamic} illustrates the low-level MDP space that comprises eight regions, along with the change of skill value functions with the execution of skills. Four skills are trained to reach colored regions: red $o^r$, yellow $o^y$, black $o^b$, and grey $o^g$, which are associated with corresponding skill value functions $V_{o^r}$, $V_{o^y}$, $V_{o^b}$, and $V_{o^g}$. The task entails first reaching the yellow goal, followed by the red one. Initially, executing one step of skill $o^y$ moves the agent into the yellow region, resulting in an increase in $V_{o^y}$ until a predetermined threshold is achieved. Subsequently, skill $o^r$ directs the agent to the red region, leading to an increase in $V_{o^r}$. The above process shows that high values in the VFS indicate a successful completion of the task in the original state space.

\begin{figure}[htbp!]
    \centering
    \begin{center}
    \includegraphics[width=0.9\columnwidth]{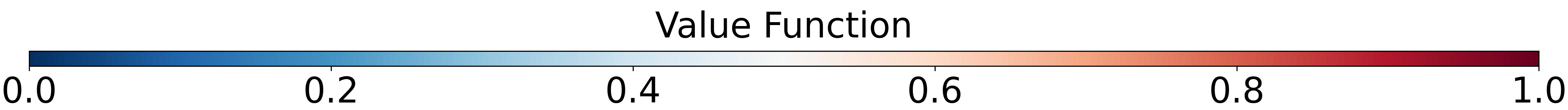}
    \end{center}
    \vspace{-15pt}  
    
    \hspace*{-0.5em}
    \subfloat{\includegraphics[width=0.45\columnwidth]{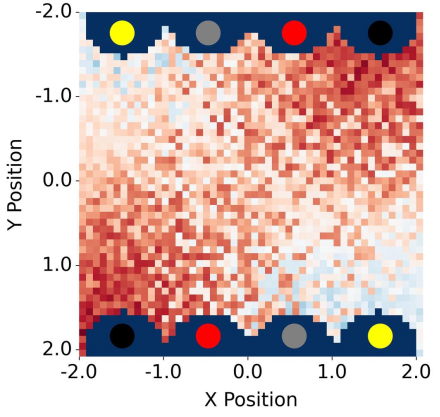}}%
    \hspace{-0.5em}
    \subfloat{\includegraphics[width=0.45\columnwidth]{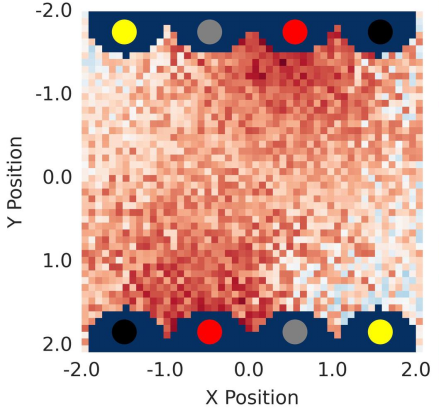}}%
    \hspace*{-0.5em}%
    
    \vspace{-12pt}  
    
    \hspace*{-0.5em}%
    \subfloat{\includegraphics[width=0.45\columnwidth]{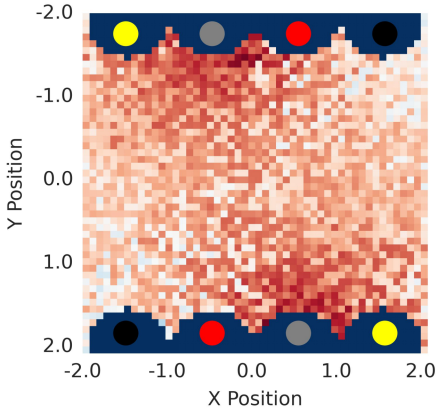}}%
    \hspace{-0.5em}%
    \subfloat{\includegraphics[width=0.45\columnwidth]{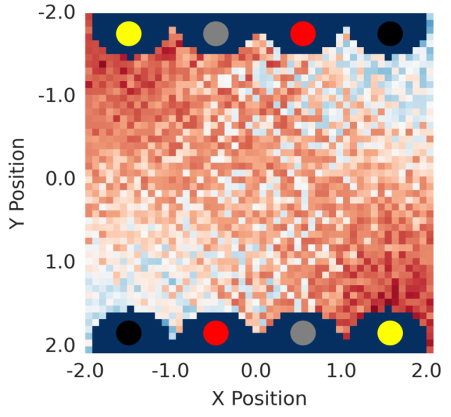}}%
    \hspace*{-0.5em}%
    
    \vspace{-6pt}  
    \caption{\footnotesize Value function visualized in discrete grids ($o^b$ upper left, $o^r$ upper right, $o^w$ lower left and $o^y$ lower right subfigure). Colorbar: value scale. Grid: designated location of the agent and its value function.}
    \label{fig:safety-gym vfs}
\end{figure}

Fig.\ref{fig:safety-gym vfs} displays skill value functions for four policies trained within the same environment. Eight colored dots mark different zones. Each grid represent a discretized $(x, y)$ coordinates and a fixed orientation of $\theta = 0$. High intensity in skill value functions is concentrated around target regions, while lower values are observed in areas further away from these targets.

\vspace{-10pt}
\vspace{-20pt}

\subsection{Planning in Value Function Space}\label{subsec:planning}
\vspace{-2pt}

We first define the \textbf{reach predicate} for STL as $\mu = V_{o^i} - \epsilon_{\text{reach}}$, where $o^i$ is the high-level policy to achieve goal $g_i$ and $\epsilon_{\text{reach}} \in [0, 1]$. If $V_{o^i} > \epsilon_{\text{reach}}$, we interpret that $g_i$ is fulfilled. In addition, we define the \textbf{avoid predicate} as $\mu = \epsilon_{\text{avoid}} - V_{o^j}$ where $o^j$ is the high-level policy for avoiding the goal $g_j$ and $\epsilon_{\text{avoid}} \in [0, 1]$. If $V_{o^j} < \epsilon_{\text{avoid}}$, we interpret that $g_j$ is successfully avoided at the current step. These predicates are utilized to construct the given STL task in this work. 

Given the STL task $\phi$ and the learned transition dynamics $\hat{f}_\theta$ of VFS, we recast \textbf{Problem \ref{prob:skill_optimization}} into an optimization problem aimed at synthesizing an optimal skill sequence $(o^*_0, o^*_1, ..., o^*_{\bar{T}-1})$ that maximizes the robustness score of $\phi$. 

\vspace{-10pt}
\begin{equation}
    o^*_0, o^*_1, ..., o^*_{\bar{T}-1} = \argmax_{o_0, o_1, ..., o_{\bar{T}-1}} \rho(\hat{Z}_{0:\bar{T}}, \phi)
    \vspace{-5pt}
\end{equation}
where $\hat{Z}_0 = Z(s_0)$ and $\hat{Z}_{\bar{t}+1} = \hat{f}_\theta(\hat{Z}_{\bar{t}}, o^i)$. $\hat{Z}_{0:\bar{T}}$ is the trajectory generated by the skill sequence $(o^*_0, o^*_1, ..., o^*_{\bar{T}-1})$.

Then we propose the STL guided skill planning (STLSP) algorithm. Starting from the initial MDP state $s_0$, we obtain the initial VFS state $Z_0 = Z(s_0)$. Next in Line 1 of Algorithm~\ref{alg:stlvfs}, we utilize random shooting method~\cite{bertsekas2025course} to generate a batch $B$ of potential skill sequences. For each candidate $b \in B$, we compute the corresponding trajectory $\hat Z_{0:\bar{T}}$ using the learned dynamics $\hat{f}_\theta$ in the VFS, resulting in $|B|$ candidate skill sequences, as detailed in Line 2. Then we evaluate the robustness score $\rho(\hat{Z}_{0:\bar{T}}, \phi)$ for each sequence in Line 3. Ultimately, the skill sequence with the highest score is executed, where each skill is repeated for $\tau$ steps in the original MDP state space.

\vspace{-8pt}
\begin{algorithm}[ht!]
\caption{STLSP}
\label{alg:stlvfs}
\begin{algorithmic}[1]
\REQUIRE STL task $\phi$, a set of skills $O$, initial state $s_0$, VFS state $Z_0 = Z(s_0)$, VFS dynamics $\hat{f}_\theta$
\ENSURE Skill sequence $(o^*_0, o^*_1, ..., o^*_{\bar{T}-1})$

\STATE Generate batch $B \sim \mathcal{U}(O_{0:\bar{T}})$;
\STATE Simulate trajectories:
\vspace{-5pt}
\begin{equation*}
\begin{aligned}
    \hat{Z}_{0:\bar{T}} &= (\hat{Z}_0, \hat{f}_\theta(\hat{Z}_0, o_1), \dots, \hat{f}_\theta(\hat{Z}_{\bar{T}-1}, o_{\bar{T}-1})) \\
    \forall o^{(k)}_{0:\bar{T}-1} &\in B,\ \bar{t} \in \{0,...,\bar{T}-1\}
\end{aligned}
\vspace{-5pt}
\end{equation*}

\STATE Compute robustness scores:
\vspace{-5pt}
\begin{equation*}
    \rho_k = \rho(\hat{Z}_{0:\bar{T}}^{(k)}, \phi),\ \forall k \in \{1,...,|B|\}
\end{equation*}

\STATE Return $o^*_{0:\bar{T}-1} = \underset{B}{\text{argmax}}\ \rho_k$.
\end{algorithmic}
\end{algorithm}

\vspace{-8pt}
\vspace{-5pt}
\section{Experiments}\label{sec:experiment}
\subsection{Zone Navigation}
\vspace{-2pt}

\textbf{Environment setup:} We evaluate our approach in a Safety Gym environment ZoneEnv~\cite{qiu2024instructing}. As illustrated in Fig.\ref{fig:VFDynamic}, the environment consists of 4 colors and 8 zones. The initial positions of the robot and the zones are randomly generated. We use lidar observation to observe zone objects. The lidar loops over all objects in a scene, then fills the appropriate lidar bins with the right value. The number of bins is set to 10. For 4 zones, we have 40 dimensional lidar observation space and 5 dimension of the agent state information. 

\textbf{Algorithm Implementation:}
The horizon of high-level planning in VFS is $\bar T = 24$ where each time step corresponds to the execution of a skill for $\tau = 100$ steps in the low-level environment. The threshold for reaching a goal is $\epsilon_{\text{reach}} = 0.9$, and the threshold for avoiding a zone is $\epsilon_{\text{avoid}} = 0.2$. The batch size of random shooting is $|B| = 10000$. The VFS dynamic neural network features two hidden layers, each with $1024$ units and ReLU activation for the intermediate layers. During trainings, we collect $|D| = 40000$ steps of VFS transitions and the STL robustness is assessed by toolbox \text{stl\_core\_lib}~\cite{meng2023signal}.

To assess the impact of hyper parameters on STL predicates in the VFS, we evaluated fixed thresholds $\epsilon_{\text{reach}}$ and $\epsilon_{\text{avoid}}$ to reach and avoid zones. $\phi_1$ from Equation (\ref{eq:3_tasks}) encompasses both reachability and avoidance tasks and we measured the Success Rate (SR) and Number of Collisions (NoC). We conducted ten experiments for each combination of thresholds, randomly vary the positions of the agent and zone, and document the results in Table~\ref{tab:epsperformence}. The combination of $\epsilon_{\text{reach}} = 0.9$ and $\epsilon_{\text{avoid}} = 0.2$ yielded a SR of $0.9$ with the lowest collisions, which is the basis for subsequent evaluations.

\vspace{-3pt}
\begin{table}[ht!]
    \centering
    \vspace{-5pt}
    \begin{tabular}{llcc}
    \toprule
    \textbf{$\epsilon_{reach}$} & \textbf{$\epsilon_{avoid}$} & SR & NoC \\
    \midrule
    0.7 & 0.3 & 0.8 & 193.3 \\
    0.7 & 0.2 & 0.8 & 55.9 \\
    0.7 & 0.1 & 0.7 & 28.0 \\
    0.8 & 0.3 & 0.9 & 128.2 \\
    0.8 & 0.2 & 0.8 & 94.2 \\
    0.8 & 0.1 & 0.7 & 54.3 \\
    0.9 & 0.3 & 0.9 & 120.8 \\
    0.9 & 0.2 & 0.9 & 53.4 \\
    0.9 & 0.1 & 0.8 & 19.5 \\
    \bottomrule
    \end{tabular}
    \caption{\footnotesize Hyper-parameter of VFS predicates $\epsilon_{reach}$ and $\epsilon_{avoid}$}
    \label{tab:epsperformence}
    \vspace{-18pt}
\end{table}

\begin{figure}[!htbp]
\centering
    \subfloat[\footnotesize Reach $\&$ Stay]{\includegraphics[width=0.33\columnwidth]{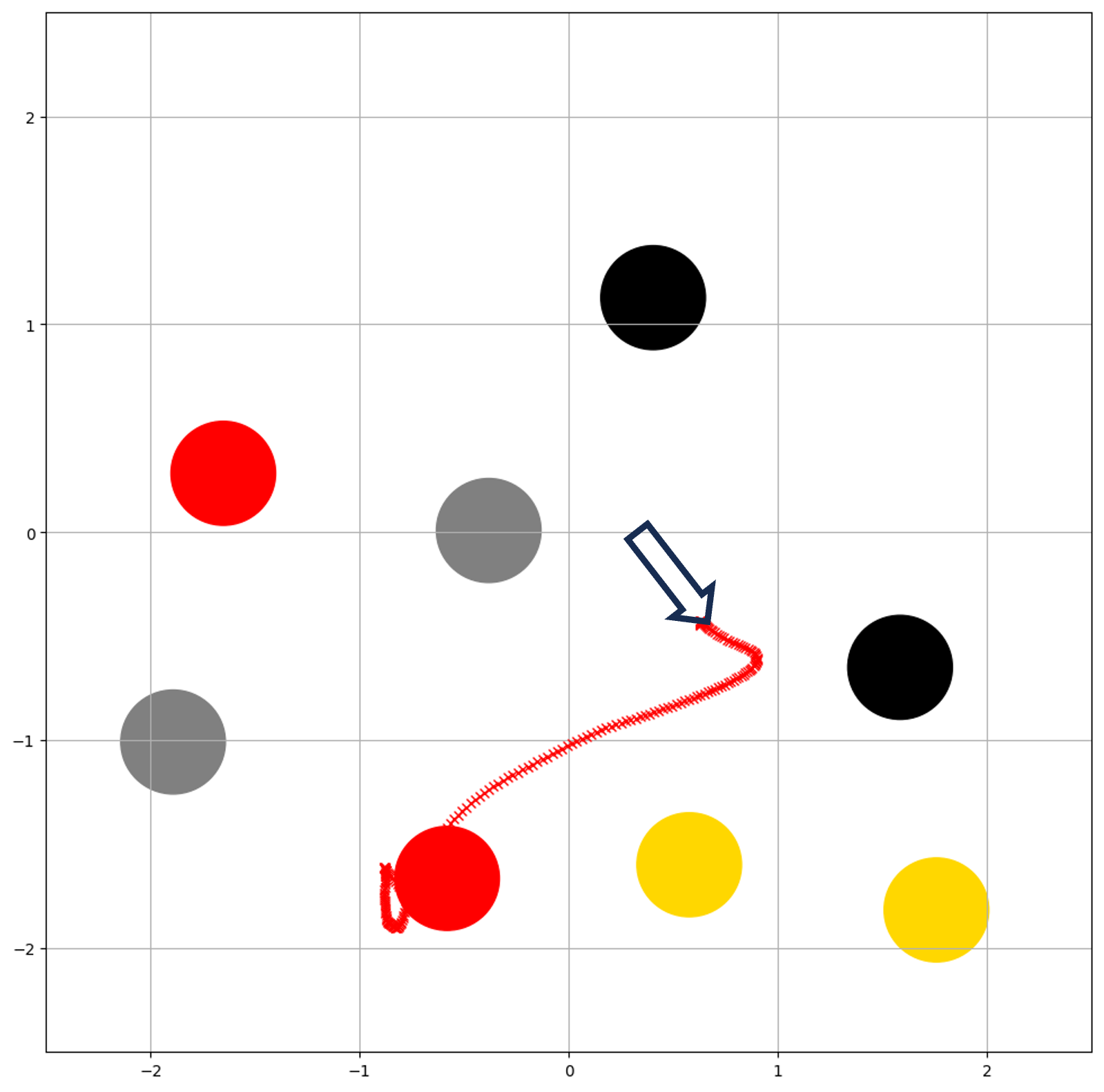}}%
    \subfloat[\footnotesize Reach Avoid]{\includegraphics[width=0.33\columnwidth]{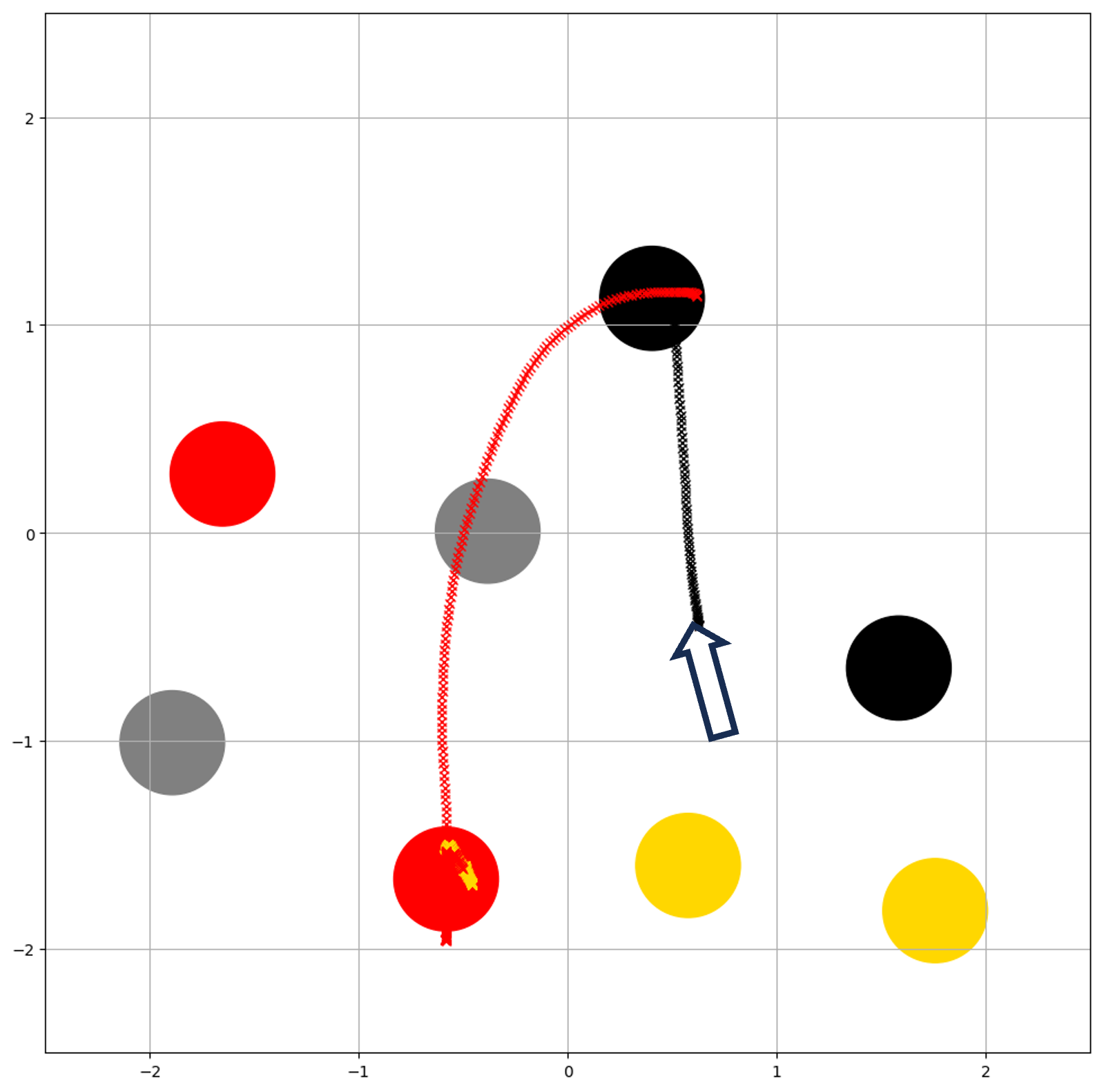}}%
    \subfloat[\footnotesize Sequential Reach]{\includegraphics[width=0.33\columnwidth]{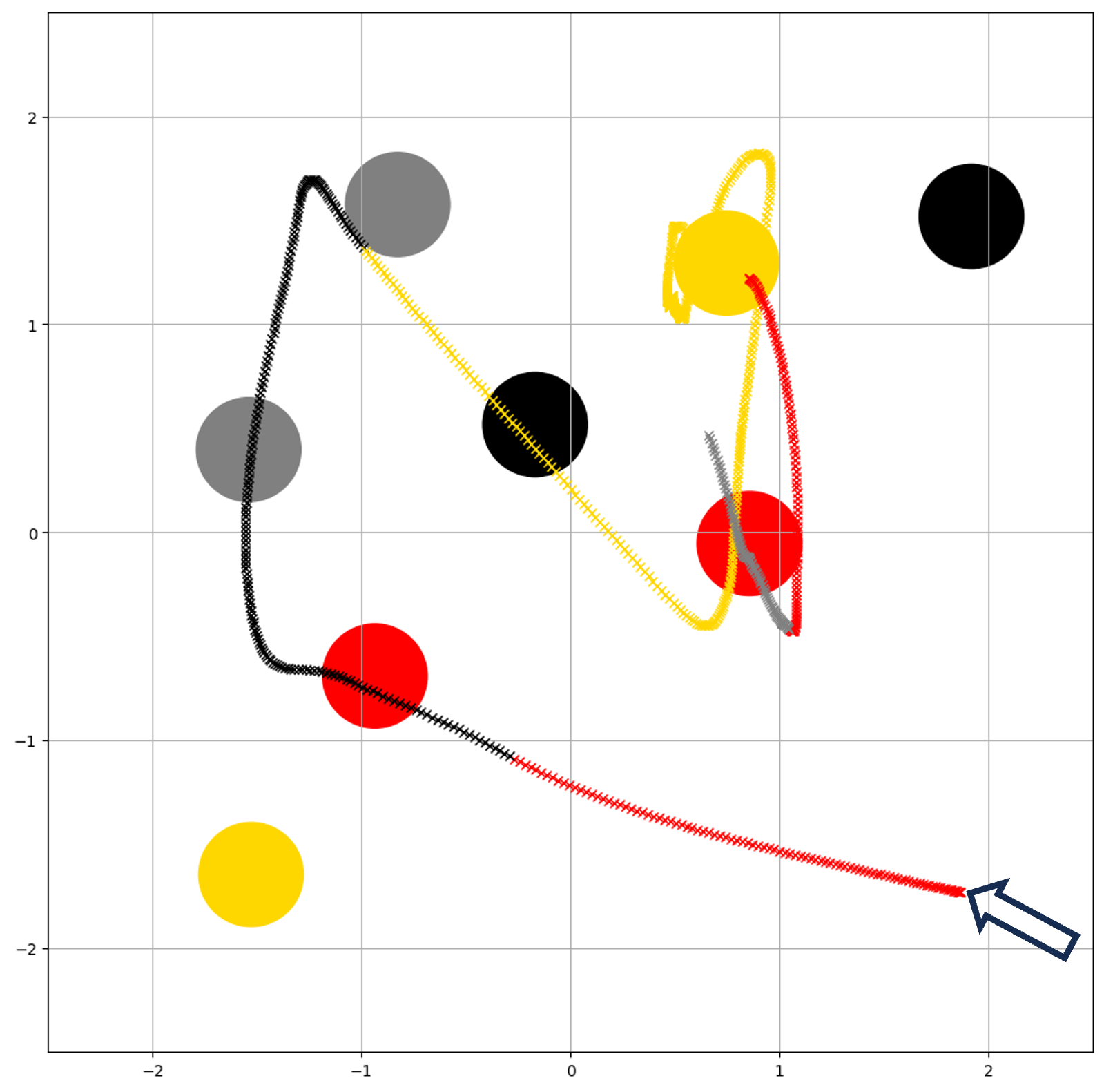}}%
    
    \caption{\footnotesize Trajectories generated by STLSP for tasks in Equations (\ref{eq:case_study}). Arrows: initial positions; colored segment line: execution of a skill.}
    \label{fig:STL_tasks}
\end{figure}

\textbf{Case Study:} we consider three common tasks in Figure~\ref{fig:STL_tasks} for the study. The corresponding STL formulas are given as:

\vspace{-12pt}
\begin{equation}
    \begin{aligned}
        \phi_a &= \mathbf{F}_{[0,2]} \mathbf{G}_{[0, 5]} (V_{o^r}> \epsilon_{reach}) \\
        \phi_b &= (V_{o^y}<\epsilon_{avoid})\mathbf{U}_{[0,2]}(V_{o^r}>\epsilon_{reach} \\
               &\wedge ( V_{o^g}<\epsilon_{avoid} \mathbf{U}_{[0,2]} V_{o^b}>\epsilon_{reach})) \\
        \phi_c &= (\mathbf{F}_{[0,3]}(V_{o^r}>\epsilon_{reach} \wedge \mathbf{F}_{[0,3]}(V_{o^b}>\epsilon_{reach} \\
        &\wedge \mathbf{F}_{[0,4]} V_{o^y}>\epsilon_{reach}))
    \end{aligned}
    \label{eq:case_study}
\end{equation}
\vspace{-5pt}

In $\phi_a$, the task is to ``reach the red goal in 2 steps and stay for 5 steps". Fig.\ref{fig:STL_tasks}(a) illustrates the execution of policy $o^r$ to maintain proximity to the red goal. In $\phi_b$, the objective is to ``avoid the grey goal until reaching the black goal within 2 steps, then reach the red goal within 2 steps while always avoiding the yellow goal". This scenario is depicted in Fig.\ref{fig:STL_tasks}(b), where the black trajectory segment corresponds to skill $o^b$, followed by $o^r$. In $\phi_c$, sequential reach entails ``sequentially reaching the red goal within 3 steps, followed by the black and yellow goals". We execute $o^r$, $o^b$, and $o^y$ in order, as shown in Fig.\ref{fig:STL_tasks}(c), before adopting a random policy upon completing $\phi_c$. These examples collectively validate that our planning method successfully achieve the STL tasks.

\vspace{-18pt}
\vspace{-8pt}

\textbf{Baselines:}
For a fair comparison, we outline the capabilities of different methods in the Table~\ref{tab:methods_capability}. ``\textbf{Unknown Dynamic (UD)}'' refers to the lack of agent dynamic models. Optimization-based methods like MILP and NLP requires the model to formulate the objective function, while sampling-based methods like STLCG require a differentiable dynamic model. ``\textbf{Partial Observation (PO)}'' pertains to the insufficient knowledge of the target region needed to create the STL formula as an objective function. 
In our environment, zone objects are observed via lidar, which does not directly access their positions. Optimization-based methods, such as MILP and NLP, require full observation, including target region information. Data-driven approaches like STL RL and STLCG need this information during training but not inference. ``\textbf{Zero-shot (ZS)}'' refers to the capacity to adapt to new STL tasks without further interaction with the environment for retraining. Data-driven methods, such as STLCG and STL RL, require additional interaction for new specifications.

Methods with all three capabilities are evaluated. LTL2Action~\cite{vaezipoor2021ltl2action} leverages LTL syntax and semantics to learn task-conditioned policies that generalize to new tasks. Although STL is not involved, it exhibits zero-shot performance on unseen LTL formulas. LTL progression is disabled during inference, as it requires a labeling function, which is not available in a ``partially observable" environment. 

\vspace{-8pt}
\begin{table}[h!]
\centering
\begin{tabular}{lccc}
\hline
 & UD & PO & ZS \\ \hline
MILP~\cite{sun2022multi} & $\times$ & $\times$ & $\checkmark$ \\ 
NLP~\cite{haghighi2019control} & $\times$ & $\times$ & $\checkmark$ \\
STLCG~\cite{leung2023backpropagation} & $\times$ & $\checkmark$ & $\times$ \\ 
STL RL~\cite{singh2023stl} & $\checkmark$ & $\checkmark$ & $\times$ \\ 
LTL2Action~\cite{vaezipoor2021ltl2action} & $\checkmark$ & $\checkmark$ & $\checkmark$ \\ 
VFS~\cite{alexander2022value} & $\checkmark$ & $\checkmark$ & $\checkmark$ \\ 
Ours & $\checkmark$ & $\checkmark$ & $\checkmark$ \\ \hline
\end{tabular}
\caption{\footnotesize Baseline methods capabilities.}
\label{tab:methods_capability}
\end{table}
\vspace{-15pt}

We also implemented VFS~\cite{alexander2022value} in Safety Gym, using a task-specific trajectory $\tilde{Z}_{0:\bar T}$ as a reference. For example, the task of reaching a red zone followed by a yellow zone is encoded as $[[0.9, 0, 0, 0], \ldots, [0, 0.9, 0, 0], \ldots]$. Each vector corresponds to the value functions of the skills $o^r, o^y, o^b, o^g$ in that order. The objective is to find an optimal skill sequence $o^*_0, o^*_1, \ldots, o^*_{\bar T-1}$ that minimizes the mean squared error between the predicted trajectory $\hat{Z}_{0:\bar T}$ and the reference $\tilde{Z}_{0:\bar T}$, focusing on relevant non-zero entries. Since these methods do not involve STL tasks, we consider SR and NoC rather than STL robustness. Our methods are run across three tasks where $\text{REACH}_i$ is $V_{o^i} \geq \epsilon_{reach}$ and $\text{AVOID}_j$ is $V_{o^j} \leq \epsilon_{avoid}$:

\vspace{-10pt}
\begin{equation}\label{eq:3_tasks}
    \begin{aligned}
    \phi_1 = &\mathbf{F}_{[0, \bar{T}]} \text{REACH}_i \land \mathbf{G}_{[0, \bar{T}]} \text{AVOID}_j, \\
    \phi_2 = &\mathbf{F}_{[0, \bar{t_1}]} \text{REACH}_i \land \mathbf{F}_{[\bar{t_1}, \bar{T}]} \text{REACH}_j \land \mathbf{G}_{[0, \bar{T}]} \text{AVOID}_k, \\
    \phi_3 =&\mathbf{F}_{[0, \bar{t_1}]} \text{REACH}_i \land \mathbf{F}_{[\bar{t_1}, \bar{t_2}]} \text{REACH}_j \land \mathbf{F}_{[\bar{t_2}, \bar{T}]} \text{REACH}_k.
    \end{aligned}
\end{equation}

Key metrics are evaluated, with $\uparrow$ and $\downarrow$ indicating preferred increase and decrease, respectively:

\begin{itemize}
    \item \textbf{SR $\uparrow$:} Indicates whether the trajectory successfully completes the reach tasks in the correct order.
    \item \textbf{NoC $\downarrow$:} The number of time steps during which the agent collides with zones that should be avoided.
\end{itemize}

Since LTL2Action does not utilize STL to define tasks, we employ LTL formulas that closely resemble STL tasks for it. For VFS, we manually configure the reference trajectory $\tilde{Z}_{0:\bar{T}}$. The Table~\ref{tab:baseline_comparsion} presents the quantitative results. For each task, we conducted 100 experiments, averaging the results over these trials. Each experiment involves random initialization of the agent's position and random placement of colored zone regions. Zones $i, j, k$ are randomly selected from four colored zones for each experiment. Quantitative evaluations across three tasks $\phi_1, \phi_2, \phi_3$ demonstrate consistent performance improvement of our method over baselines. Compared with VFS, our method achieves higher SR in all tasks: 6.25\% higher in $\phi_1$, 9.52\% higher in $\phi_2$, and 11.9\% higher in $\phi_3$. Our method also reduces collisions by 44.4\% in $\phi_1$ and 9.08\% in $\phi_2$ compared to VFS. LTL2Action, trained on $\phi_1, \phi_2$ with 0.37 SR and 0.13 SR, respectively, but demonstrates limited generalization to out-of-distribution tasks $\phi_3$, and achieves SR of 0.05. While LTL2Action reports fewer collisions in $\phi_1$ and $\phi_2$, this is compromised by minimal task SR. Overall, our method maintains a desirable balance between SR and Noc.

\vspace{-8pt}
\begin{table}[h!]
\centering
\begin{tabular}{lccc}
\hline
\textbf{Methods} & \textbf{Task} & \textbf{SR $\uparrow$} & \textbf{NoC $\downarrow$} \\ 
\hline
LTL2Action & \multirow{3}{*}{$\phi_1$} & 0.37 & \textbf{42.2} \\ 
VFS      & & 0.80 & 127.74\\ 
STLSP(Ours)     & & \textbf{0.85} & 70.96\\ 
\hline
LTL2Action & \multirow{3}{*}{$\phi_2$} & 0.13 & \textbf{29.1}\\ 
VFS      &  & 0.63 & 158.89\\ 
STLSP(Ours)      &  & \textbf{0.69} & 85.39\\ 
\hline
LTL2Action & \multirow{3}{*}{$\phi_3$} & 0.05& - \\ 
VFS      &  & 0.42 & - \\ 
STLSP(Ours)      &  & \textbf{0.47} & - \\ 
\hline
\end{tabular}
\caption{\footnotesize Comparisons between our method and baselines.}
\label{tab:baseline_comparsion}
\end{table}

\vspace{-35pt}
\subsection{Robot Manipulation}
\vspace{-3pt}

\begin{figure}[h]
  \centering
  \includegraphics[width=\linewidth]{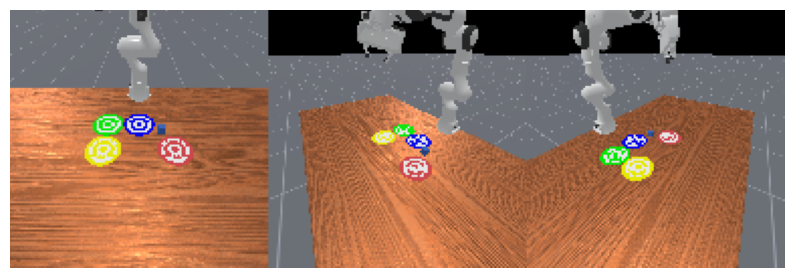}
  \vspace{-20pt}
  \caption{\footnotesize The robot manipulation environment. The tasks are pushing the cube to desired regions in order and avoid undesired regions.}
  \label{fig:robo_arm}
\end{figure}

To demonstrate the scalability of our method for more complex systems, we evaluate the performance of STLSP in a sophisticated image-based task within a manipulation setting in Figure~\ref{fig:robo_arm}. We customized the Maniskill environment~\cite{gu2023maniskill2}, where the robot relies solely on high-dimensional visual inputs and proprioceptive states. The task involves a two-finger gripper arm manipulating a cube on a table surface, with goal regions marked by red, green, yellow, and blue circular targets. In this context, $\text{REACH}_i$ indicates pushing the cube to goal region $i$, while $\text{AVOID}_j$ requires avoiding pushing the cube into region $j$, where $i$ and $j$ correspond to the colored regions. We conducted 100 experiments for each task, with cube positions and goal region positions randomly initialized for each trial. Our method achieved a SR of 65\% in task $\phi_1$, 33\% in task $\phi_2$, and 18\% in task $\phi_3$.

\vspace{-10pt}
\vspace{-15pt}
\vspace{-8pt}
\section{Conclusions}\label{sec:conclusion}
\vspace{-2pt}

This work presents a hierarchical planning framework tailored for Signal Temporal Logic tasks. We first construct a value function space (VFS) to abstract the original state space. Next, we reformulate the planning problem in VFS where reusing skills facilitates training solely for reachability policies to fulfill both reachability and safety predicates. Then we synthesize optimal skill sequences that reliably guide agents to accomplish the tasks in the original environment. Simulations validate the soundness and effectiveness of our approach, which outperforms the baselines. Looking ahead, we plan to expand our framework to accommodate more complicated planning scenarios with a large volume of skills.

\bibliographystyle{ieeetr}
\bibliography{ref}

\end{document}